\documentclass{article}

\usepackage[final,nonatbib]{neurips_2020}
\usepackage[square,numbers]{natbib}
\usepackage[utf8]{inputenc} 
\usepackage[T1]{fontenc}    
\usepackage{hyperref}       
\usepackage{url}            
\usepackage{booktabs}       
\usepackage{nicefrac}       
\usepackage{microtype}     
\usepackage{amsfonts}       
\usepackage{nicefrac}       
\usepackage{microtype}  
\usepackage{amsmath}
\usepackage{amssymb}
\usepackage{amsthm}
\usepackage{graphicx}
\usepackage{float}
\usepackage[normalem]{ulem}
\useunder{\uline}{\ul}{}
\usepackage{subcaption}
\usepackage{titlesec}
\usepackage{lastpage}
\usepackage{algorithm} 
\usepackage[noend]{algpseudocode} 

\newcommand{\KL}{D_{\mathrm{KL}}}

\title{Likelihood Regret: An Out-of-Distribution Detection Score For Variational Auto-encoder}

\author{
  Zhisheng Xiao \thanks{equal contribution} \\
  Computational and Applied Mathematics\\
  University of Chicago\\
  Chicago, IL, 60637\\
  \texttt{zxiao@uchicago.edu} \\
   \And
  Qing Yan \footnotemark[1]\\
  Department of Statistics\\
  University of Chicago\\
  Chicago, IL, 60637\\
  \texttt{yanq@uchicago.edu} \\
    \And
  Yali Amit \\
  Department of Statistics\\
  University of Chicago\\
  Chicago, IL, 60637\\
  \texttt{amit@marx.uchicago.edu} \\
}

\begin{document}

\maketitle

\begin{abstract}
    Deep probabilistic generative models enable modeling the likelihoods of very high dimensional data. An important application of generative modeling should be the ability to detect out-of-distribution (OOD) samples by setting a threshold on the likelihood. However, some recent studies show that probabilistic generative models can, in some cases, assign higher likelihoods on certain types of OOD samples, making the OOD detection rules based on likelihood threshold problematic. To address this issue, several OOD detection methods have been proposed for deep generative models. In this paper, we make the observation that many of these methods fail when applied to generative models based on  Variational Auto-encoders (VAE). As an alternative, we propose Likelihood Regret, an efficient OOD score for VAEs. We benchmark our proposed method over existing approaches, and empirical results suggest that our method obtains the best overall OOD detection performances when applied to VAEs. 
\end{abstract}

\section{Introduction}\label{intro}
In order to make reliable and safe decisions, deep learning models that are deployed for real life applications need to be able to identify whether the input data is anomalous or significantly different from the training data. Such data are called out-of-distribution (OOD) data. However, it is known that neural network classifiers can over-confidently classify OOD data into one of the training categories \cite{fool}. This observation poses a great challenge to the reliability and safety of AI \cite{safety}, making OOD detection a problem of primary importance. Several approaches have been proposed to detect OOD data based on deep classifiers \cite{baseline,simple,confidence,exposure}. Unfortunately, these methods cannot be applied to OOD detection for models trained without supervision, such as many generative models. An
appealing OOD detection approach that may work for probabilistic generative models is to use their likelihood estimates. Such models can evaluate the likelihood of input data, and if a generative model fits the training data distribution well enough, it should assign high likelihood to samples from the training distribution and low likelihood to OOD samples. 

Recent advances in deep probabilistic generative models \cite{vae, pixelcnn, pixelcnn++, glow} make generative modeling of very high dimensional and complicated data such as natural images, sequences \cite{wavenet} and graphs \cite{graphvae} possible. These models can evaluate the likelihood of input data easily and generate realistic samples, indicating that they succesfully approximate the distribution of training data. Therefore, it would appear promising to use deep generative models to detect OOD data \cite{anomaly_time}. However, some recent studies \cite{nalisnick, waic} reveal a counter intuitive phenomenon that challenges the validity of unsupervised OOD detection using generative models. They observe that likelihoods obtained from current state-of-the-art deep probabilistic generative models fail to distinguish between training data and some obvious OOD input types that are easily recognizable by humans. For example, \cite{nalisnick} shows that generative models trained on CIFAR-10 output higher likelihood on SVHN than on CIFAR-10 itself, despite the fact that images in CIFAR-10 (contains dogs, trucks, horses, etc.) and SVHN (contains house numbers) have very different semantic content. 

At this point, no effective method has been discovered to ensure these generative models make the correct likelihood assignment on OOD data. Alternatively, some new scores based on likelihood are proposed to alleviate this issue \cite{ratio, typicality, complexity}. The OOD detection is performed by setting thresholds on the new scores rather than on likelihood.  Some of these methods obtain impressive OOD detection performance on invertible flow-based models \cite{glow} and auto-regressive models \cite{pixelcnn++}. Interestingly, we observe that these scores can be much less effective for Variational Auto-encoders (VAE), an important type of probabilistic generative models. The failure of current OOD scores on VAE suggests that a new score is necessary. To this end, in this paper we propose a simple yet effective metric called Likelihood Regret (LR) to detect OOD samples with VAEs. The Likelihood Regret of a single input can be interpreted as the log ratio between its likelihood obtained by the posterior distribution optimized individually for that input and the likelihood approximated by the VAE. We conduct comprehensive experiments to evaluate our proposed score on a variety of image OOD detection tasks, and we show that it obtains the best overall performance.

\vspace{-3mm}
\section{Background} \label{bg}
\subsection{Variational Auto-Encoder}
VAE \cite{vae, rezende} is an important type of deep probabilistic generative model with many practical applications \cite{liang2018variational, liu2017unsupervised, gregor2018temporal}. It uses a latent variable $\mathbf{z}$ with prior $p(\mathbf{z})$, and a conditional distribution $p_{\theta}(\mathbf{x}|\mathbf{z})$, to model the observed variable $\mathbf{x}$. The generative model, denoted by $p_{\theta}(\mathbf{x})$, can be formulated as $p_{\theta}(\mathbf{x})=\int_{\mathcal{Z}} p_{\theta}(\mathbf{x} | \mathbf{z}) p(\mathbf{z})\mathrm{d} \mathbf{z}$. However, direct computation of this likelihood is intractable in high dimensions, so variational inference is used to derive a lower bound on the log likelihood of $\mathbf{x}$. This leads to the famous evidence lower bound (ELBO):
\begin{align}
    \log p_{\theta}(\mathbf{x}) & \ge\mathbb{E}_{q_{\phi}(\mathbf{z} | \mathbf{x})}\left[\log p_{\theta}(\mathbf{x} | \mathbf{z})\right]-\KL\left[q_{\phi}(\mathbf{z} | \mathbf{x}) \| p(\mathbf{z})\right]\notag \\ & \triangleq \mathcal{L}(\mathbf{x} ; \theta, \phi),\label{ELBO}
\end{align}
where $q_{\phi}(\mathbf{z} | \mathbf{x})$ is the variational approximation to the true posterior distribution $p_{\theta}(\mathbf{z} | \mathbf{x})$. Both $q_{\phi}(\mathbf{z} | \mathbf{x})$ and $p_{\theta}(\mathbf{x} | \mathbf{z})$ are parameterized by neural networks with parameters $\phi$ (encoder) and $\theta$ (decoder), respectively. The VAE is trained by maximizing $\mathcal{L}(\mathbf{x} ; \theta, \phi)$ over the training data. 

Unlike generative models using exact inference so that the likelihood can be directly computed, VAE only outputs a lower bound of the log likelihood and the exact log likelihood needs to be estimated, usually by an importance weighted lower bound \cite{IWAE}:
\begin{align} \label{iwae}
 \log p_{\theta}(\mathbf{x}) \geq \mathbb{E}_{\mathbf{z}^1,...,\mathbf{z}^K \sim {q_{\phi}(\mathbf{z} | \mathbf{x})}}\left[\log \frac{1}{K} \sum_{k=1}^{K} \frac{p_{\theta}\left(\mathbf{x}| \mathbf{z}^{k}\right)p(\mathbf{z}^k)}{  q_{\phi}\left(\mathbf{z}^{k} | \mathbf{x}\right) }\right] \triangleq \mathcal{L}_{K}(\mathbf{x};\theta,\phi),
\end{align}
where each $\mathbf{z}^{k}$ is a sample from the variational posterior $q_{\phi}\left(\mathbf{z} | \mathbf{x}\right)$. 

While the prior $p(\mathbf{z})$ and the variational posterior $q_{\phi}\left(\mathbf{z} | \mathbf{x}\right)$ are often chosen to be Gaussians, there are multiple choices for the decoding distribution $p_{\theta}\left(\mathbf{x} | \mathbf{z}\right)$ depending on the type of data. 
In this paper, we follow the settings of VAE experiments in \cite{nalisnick} and choose the decoding distribution to be a factorial 256-way categorical distribution (corresponding to 8-bit image data) on each pixel. Note that the same data distribution is assumed by PixelCNN \cite{pixelcnn}. 

\subsection{Problems with OOD Detection using Probabilistic Generative Models}
Suppose we have a set of $N$ training samples $\left\{\boldsymbol{\mathbf{x}}_{i}\right\}_{i=1}^{N}$ drawn from some underlying data distribution $\mathbf{x}_i \sim p(\mathbf{x})$. Our goal is to decide whether a test sample $\mathbf{x}$ is OOD, which, by definition in \cite{liang2018enhancing}, means that $\mathbf{x}$ has low density under $p(\mathbf{x})$. Probabilistic generative models $p_{\theta}(\mathbf{x})$ are trained on the set of training samples by maximizing the likelihood (or lower bound of likelihood). It is well known that maximizing the likelihood $p_{\theta}(\mathbf{x})$ is equivalent to minimizing $\KL\left[p(\mathbf{x}) \| p_{\theta}(\mathbf{x})\right]$, and thus a well trained generative model provides a good approximation to the true data distribution $p(\mathbf{x})$. Therefore, OOD data should have low likelihood under $p_{\theta}(\mathbf{x})$, since they stay in low probability regions of $p(\mathbf{x})$.

\begin{figure}[ht]
    \centering
    \begin{subfigure}{.35\linewidth}
        \includegraphics[scale=0.35]{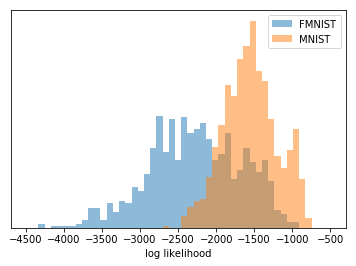}
        \caption{}
    \end{subfigure}
    \hskip2em
   \begin{subfigure}{.35\linewidth}
        \includegraphics[scale=0.35]{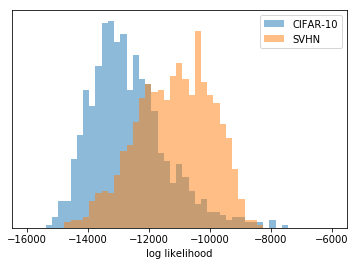}
        \caption{}
    \end{subfigure}
    \caption{\label{fig:hist1}
     Histogram that compares the log likelihood of test samples from \textbf{(a)}:  Fsahion MNIST and MNIST on a VAE trained on Fashion MNIST, and \textbf{(b)}: CIFAR-10 and SVHN on a VAE trained on CIFAR-10. Both experiments show that VAEs may assign high likelihoods to OOD samples.}
     
\end{figure}

\begin{figure}[ht]
    \centering
    \begin{subfigure}{.45\linewidth}
        \includegraphics[scale=0.5]{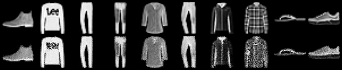}
        \caption{}
    \end{subfigure}
    \hskip2em
   \begin{subfigure}{.45\linewidth}
        \includegraphics[scale=0.5]{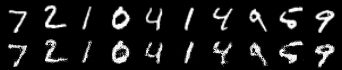}
        \caption{}
    \end{subfigure}
    \hskip2em
   \begin{subfigure}{.45\linewidth}
        \includegraphics[scale=0.5]{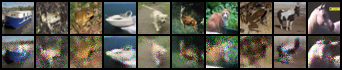}
        \caption{}
    \end{subfigure}
    \hskip2em
   \begin{subfigure}{.45\linewidth}
        \includegraphics[scale=0.5]{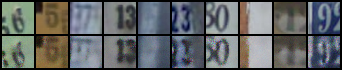}
        \caption{}
    \end{subfigure}
    \caption{\label{fig:recon}
     For each subfigure, the top row contains original images and the bottom row contains the reconstructed images. \textbf{(a), (b):} reconstruction of Fashion MNIST and MNIST images by a VAE trained on Fashion MNIST. \textbf{(c), (d):} reconstruction of CIFAR-10 and SVHN images by a VAE trained on CIFAR-10.}
\end{figure}

However, the above argument fails in practice, as noted in \cite{nalisnick,waic}. In particular, \cite{nalisnick} observe that almost all major types of probabilistic generative models, including VAE, flow-based model and auto-regressive model, can assign spuriously high likelihood to OOD samples. In Figure \ref{fig:hist1}, we confirm that such a likelihood misalignment does exist on VAE, which is the model we focus on in this paper. We further observe that VAEs obtain  surprisingly good reconstruction quality on OOD data (Figure \ref{fig:recon}), indicating that they model OOD samples very well. These observations suggest that VAEs do not really regard some samples from completely different distributions as OOD, and it is extremely unreliable to use likelihood as an OOD detector.  

\section{Related Work} \label{related_work}
Some specific types of generative models are shown to correctly assign low likelihoods to OOD samples. For example, \cite{ebm} show that energy-based models do not suffer seriously from the likelihood misalignment issue. However, the test likelihoods of in-distribution and OOD samples still exhibit significant overlap.  \cite{maaloe2019biva} shows that bidirectional-inference VAEs with very deep hierarchy of latent variables can slightly alleviate the likelihood misalignment issue, however, this comes at the cost of much worse likelihood estimates. 

To the best of our knowledge, there is no consistent way to train generative models that can effectively detect OOD samples only by looking at likelihood, and therefore people seek to design new OOD scores. \cite{waic} observe that OOD samples have higher variance likelihood estimates under different independently trained models. Although the metric obtained from an ensemble of models performs well, training multiple models can be computationally expensive. \cite{typicality} use an explicit
test of typicality and \cite{song2019unsupervised} propose an OOD detection score that leverages the batch normalization. However, both \cite{typicality, song2019unsupervised} can only determine whether a {\it batch} of samples is an outlier or not, which greatly limits their applications to the real OOD detection task, where normally we want to detect if a single sample is in- or out-of-distribution. 

Perhaps \cite{ratio,complexity} have the closest connection with our work. \cite{ratio} propose the use of a likelihood-ratio test by taking the ratio between the likelihood obtained from the model and from a background model which is trained on random perturbations of input data. \cite{complexity} hypothesizes that the likelihood of generative models are biased by the complexity of the inputs, and they offset the bias by a factor that measures the input complexity. They use the length of lossless compression of the image as the complexity factor, and their OOD score can also be interpreted as a likelihood-ratio test statistic by regarding the compressor as a universal model. \cite{ratio,complexity} obtain great OOD detection performances with Glow and Pixel-CNN, however, neither of them evaluates their methods on VAE. Later in this paper, we will show that their OOD scores do not work well with VAEs, suggesting the need to design an effective OOD score for VAEs. 

Previously, Auto-encoders and VAEs were widely used for anomaly detection \cite{vae_anomaly}, and empirical success was achieved in web applications \cite{kpi} and time series \cite{time}. However, their main ideas are based on the hope that VAEs cannot reconstruct OOD samples well \cite{akcay2018ganomaly,zong2018deep,dehaene2020iterative}, which was later proven to be false in many cases. \cite{denouden2018improving} incorporates both reconstruction loss and the Mahalanobis distance \cite{lee2018simple} in the latent space as an OOD detection score. Their method improves the performance of reconstruction based OOD detection, but we will show that it is still not effective in many experiments. \cite{csiszarik2019negative} propose a way to enable OOD detection with VAE, but they require training with negative samples.

\section{Likelihood Regret for OOD Detection using VAEs}
\subsection{Why We Need a New OOD Score for VAE?} \label{necessity}
Before introducing our method, we would like to emphasize why it is necessary to design a metric of OOD detection for VAE. One might ask, given that OOD scores like \cite{ratio,complexity} work so well for Glow and PixelCNN, why not just apply them to VAEs? We point out a key difference between VAE and other generative models in Table \ref{table:bpd1}, where we trained different generative models on Fashion MNIST and CIFAR-10 and report the test bits-per-dimension (BPD) of different datasets. The BPD is computed by normalizing the negative log likelihood by the dimension of an input: $\text{BPD}(\mathbf{x}) = \frac{-\log p_{\theta}(\mathbf{x})}{\log(2)\cdot d}$. We observe that while all generative models exhibit similar behavior of assigning high likelihoods to certain types of OOD samples, the relative change in average likelihood across different datasets are different. In particular, the average test likelihoods of VAE across different datasets have a much smaller range than that of Glow and PixelCNN, suggesting that the likelihoods of in-distribution and OOD samples are much less ``separated away" in VAE. The reason is probably that flow-based and auto-regressive models try to model each pixel of the input image, while the bottleneck structure in VAE forces the model to ignore some information.

Less separated likelihoods of in-distribution and OOD samples make OOD detection for VAE harder, as some OOD scores rely on the gap of likelihoods. We will empirically show in Section \ref{exp} that current state-of-the-art generative model OOD scores are much less effective for VAE, partly due to the smaller gap of likelihoods. This suggests the need for a new OOD score for VAE.  

\begin{table}[]   
\centering  
\begin{subtable}{.46\textwidth}
\centering
   \begin{tabular}{llll}
   \toprule
   & VAE  & Glow & PixelCNN \\
   \midrule
   FMNIST   & 3.20  & 3.25 & 2.68     \\
   MNIST    & 2.18 & 2.10  & 1.51      \\
   Constant & 4.21 & 1.35 & 4.01     \\
   Noise    & 8.89 & 8.96 & 7.71    \\
   \bottomrule
   \end{tabular}
\caption{Trained on Fashion MNIST}
\end{subtable}
\begin{subtable}{.46\textwidth}
\centering 
  
   \begin{tabular}{llll}
   \toprule
   & VAE  & Glow & PixelCNN \\
   \midrule
   CIFAR-10    & 4.12 & 4.05 & 3.57     \\
   SVHN     & 3.63 & 2.65 & 2.39     \\
   Constant & 2.49 & 0.76 & 0.85     \\
   Noise    & 5.98 & 9.38 & 10.17   \\
   \bottomrule
   \end{tabular}
\caption{Trained on CIFAR-10}
\end{subtable}
\caption{Average BPD of different test datasets for VAE, Glow and PixelCNN trained on Fashion MNIST and CIFAR-10.} \label{table:bpd1}
\vspace{-4mm}
\end{table}

\begin{algorithm}
\caption{Computing Likelihood Regret (LR)}\label{compute_lr}
\begin{algorithmic}[1]
\Require Test sample $\mathbf{x}$, trained VAE $(E_{\phi^*}, D_{\theta^*})$, number of posterior samples for likelihood estimation $K$, number of optimization step $S$, learning rate $\gamma$.
\State $L_{\text{VAE}} = \mathcal{L}_K(\mathbf{x};\theta^*,\phi^*)$  \Comment{Estimate the log likelihood of $\mathbf{x}$ under the VAE model by \eqref{iwae}}
\State Set $\phi = \phi^*$ 
\For{$S$ iterations}
\State $\phi \leftarrow \text{Adam}(\phi,\nabla_{\phi}(-\mathcal{L}(\mathbf{x} ; \theta^*, \phi)),\gamma)$  \Comment{Optimize $\phi$ by maximizing the ELBO objective}
\EndFor
\State $L_{\text{OPT}} = \mathcal{L}_K(\mathbf{x};\theta^*,\phi)$  \Comment{Estimate the log likelihood of $\mathbf{x}$ with optimized encoder}
\State $\text{LR} = L_{\text{OPT}} - L_{\text{VAE}}$
\end{algorithmic}
\end{algorithm}
\subsection{Our Proposed Score}
Our proposed OOD detection score, called Likelihood Regret (LR), measures the log likelihood improvement of the model configuration that maximizes the likelihood of an individual sample over the configuration that maximizes population likelihood. Intuitively, if a generative model is well trained on the training data distribution, given an in-distribution test sample, the improvement of likelihood by replacing the current model configuration with the optimal one for the single sample should be relatively small, hence resulting in low LR. In contrast, for an OOD test sample, since the model has not seen similar samples during training, the current model configuration is much less likely to be close to the optimal one, hence the LR could be large. Therefore, LR can serve as a good OOD detection score. However, in practice, a generative model can easily overfit when being optimized on a single sample and obtain very high likelihood, thus we have to seek some form of regularization on the model configuration that is being optimized. Luckily, the bottleneck structure of VAE provides a natural regularization, by restricting the optimization to the parameters of the variational posterior distribution.

Formally, suppose we have a VAE with encoder $E_{\phi}$ and decoder $D_{\theta}$. As commonly used in VAE \cite{vae}, $q_\phi (\mathbf{z}|\mathbf{x}) \sim \mathcal{N}(\mu_{\mathbf{x}},\sigma_{\mathbf{x}}^2\mathbf{I})$, so the encoder outputs the mean and variance: $E_\phi(\mathbf{x}) = (\mu_{\mathbf{x}},\sigma_{\mathbf{x}})$. For clarity, we use $\tau({\mathbf{x}},\phi)$ to denote the sufficient statistics $(\mu_{\mathbf{x}},\sigma_{\mathbf{x}})$ of $q_\phi (\mathbf{z}|\mathbf{x})$. Further we express ELBO in \eqref{ELBO} as $\mathcal L(\mathbf{x};\theta,\tau(\mathbf{x},\phi))$ to emphasize its direct dependency on $\tau(\mathbf{x},\phi)$. During training, based on empirical risk minimization (ERM) criterion, our objective is to obtain network parameters $\Theta^* = (\phi^*,\theta^*)$ that maximize the population log likelihood $\log p_{\theta}(\mathbf{x})$.
In other words, $\Theta^*$ are the parameters that achieve best \textit{average} log likelihood over the finite training set. In practice, we train VAE by maximizing ELBO instead of log likelihood. Since ELBO is a lower bound of log likelihood, we can regard maximizing EBLO as a good surrogate for maximizing log likelihood, so
\begin{equation}
    (\phi^*,\theta^*) \approx \arg\max_{(\phi,\theta)}  \frac{1}{n}\sum_{i=1}^n \mathcal L(\mathbf{x}_i;\theta,\tau(\mathbf{x}_i,\phi)).
\end{equation}


 Since $q_\phi(\mathbf{z}|\mathbf{x})$ can be fully determined by $\tau(\mathbf{x},\phi)$, we also denote $\Theta = (\tau,\theta)$ or $\Theta = (\tau(\cdot,\phi),\theta)$ as an abuse of notation.

For a specific test input $\mathbf{x}$, 
we can fix the decoder parameters $\theta^*$, and find the optimal configuration of the variational posterior distribution parameter $\hat{\tau}(\mathbf{x}) = (\hat{\mu}_\mathbf{x},\hat{\sigma}_\mathbf{x})$ that maximizes its \textit{individual} ELBO: 
\begin{equation}\label{opt_z}
    \hat{\tau} (\mathbf{x}) = \underset{\tau}{\text{argmax}}\,  L(\mathbf{x};\theta^*,\tau).
\end{equation}
In other words, $\hat\tau(\mathbf{x})$ is the optimal posterior distribution of the latent variable $\mathbf{z}$ given the particular input $\mathbf{x}$ and the optimal decoder $\theta^*$ obtained from the training set. Now we define the Likelihood Regret (LR) of the input data $\mathbf{x}$ as
\begin{equation}\label{lik_regret}
    \text{LR}(\mathbf{x}) = L(\mathbf{x};\theta^*,\hat\tau (\mathbf x)) - L(\mathbf{x};\theta^*,\phi^*).
\end{equation}

LR can also be interpreted as the log ratio of the likelihood obtained from the generative model (VAE) with the optimal configuration of the variational posterior distribution for an individual input, to its likelihood obtained from the VAE trained on training set. This interpretation connects our method to \cite{ratio,complexity}, which also have a likelihood ratio interpretation. The naming of our method is motivated by the concept of regret in online learning, which measures how `sorry' the learner is, in retrospect, not to have followed the predictions of the optimal hypothesis \cite{shalev2007online}. 

\textbf{Implementation: }The major component of evaluating LR is computing $\hat{\tau} (\mathbf{x})$ defined in \eqref{opt_z}, namely the configuration $\tau(\mathbf{x})$ that maximizes the likelihood (ELBO) for a single sample. To do this, we fix the decoder $\theta^*$, and apply iterative optimization algorithms on $\tau$ with initialization $\tau^*(\mathbf{x}) = E_{\phi^*}(\mathbf{x})$ using $\mathcal{L}(\mathbf{x};\theta^*,\tau)$ as the objective function until convergence. As an alternative approach, instead of optimizing $\tau$ directly, we can also optimize the parameters $\phi$ (initialized as $\phi^*$) of the encoder given $\mathbf{x}$ and $\theta = \theta^*$. We summarize the computation of Likelihood Regret in Algorithm \ref{compute_lr}. 
\vspace{-3mm}
\section{Results}\label{exp}
In this section, we conduct experiments to benchmark the performance of Likelihood Regret on OOD detection tasks on image datasets. For most experiments, we train VAEs with samples only from the training set of in-distribution data (Fashion-MNIST and CIFAR-10), and use test samples from different datasets to measure the OOD performances. Details regarding datasets and experimental settings can be found in Appendix \ref{appA}.

\begin{figure}[ht]
    \centering
    \begin{subfigure}{.35\linewidth}
        \includegraphics[scale=0.35]{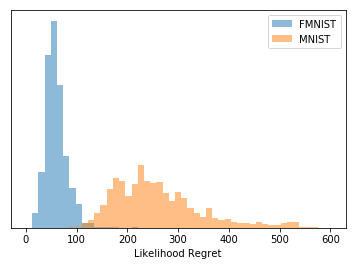}
        \caption{}
    \end{subfigure}
    \hskip2em
   \begin{subfigure}{.35\linewidth}
        \includegraphics[scale=0.35]{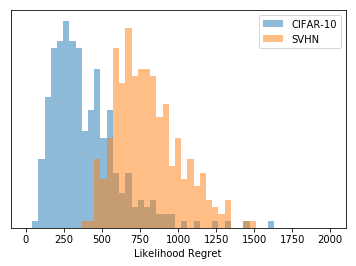}
        \caption{}
    \end{subfigure}
    \caption{\label{fig:hist2}
     Histogram that compares the Likelihood Regret of test samples from \textbf{(a)}:  Fsahion MNIST and MNIST on a VAE trained on Fashion MNIST, and \textbf{(b)}: CIFAR-10 and SVHN on a VAE trained on CIFAR-10. Both experiments show that OOD samples tend to have higher LR, as expected.}
\end{figure}

We first follow the setting of \cite{nalisnick} and conduct the following two experiments as a proof-of-concept: (a) Fashion-MNIST as in-distribution and MNIST as OOD, and (b) CIFAR-10 as in-distribution and SVHN as OOD. We train VAEs on the training set of Fashion MNIST and CIFAR-10, and compute the Likelihood Regret for $1000$ random samples from the test set of in-distribution data and corresponding OOD data. We plot the histograms of LR in Figure \ref{fig:hist2}. Comparing Figure \ref{fig:hist1} with Figure \ref{fig:hist2}, we observe that the VAE will assign higher likelihoods to OOD samples, while Likelihood Regret can largely correct such likelihood misalignment. OOD samples typically have larger LR than in-distribution samples, which confirms the effectiveness of our OOD score. 

\begin{figure}[ht]
    \centering
    \begin{subfigure}{.35\linewidth}
        \includegraphics[scale=0.35]{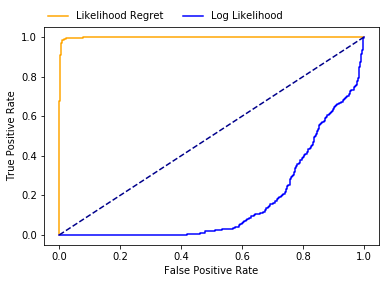}
        \caption{Fashion MNIST v.s. MNIST}
    \end{subfigure}
    \hskip2em
   \begin{subfigure}{.35\linewidth}
        \includegraphics[scale=0.35]{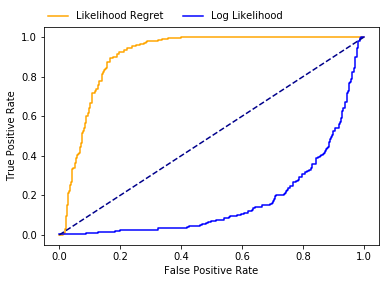}
        \caption{CIFAR-10 v.s. SVHN}
    \end{subfigure}
    \caption{\label{fig:roc1}
     Comparing the ROC curves of using Likelihood Regret and Log Likelihood for OOD detection. On Fashion MNIST v.s. MNIST experiment, Likelihood Regret improves the AUC-ROC of OOD detection from 0.165 to 0.999. On CIFAR-10 v.s. SVHN experiment, Likelihood Regret improves the AUC-ROC of OOD detection from 0.161 to 0.876. }
     \vspace{-4mm}
\end{figure}
\subsection{Metrics}
Like other OOD scores, Likelihood Regret needs a threshold when applied to the OOD detection tasks. The choice of a threshold depends on the particular application. To quantitatively evaluate our method, we mainly use Area Under the Curve-Receiver Operating Characteristics (AUCROC$\uparrow$) as a metric. AUCROC is commonly used in OOD detection tasks \cite{baseline}, and it provides an effective performance summary across different score thresholds \cite{roc,roc2}. In Figure \ref{fig:roc1}, we plot the ROC curves for the above two experiments, and results show that log likelihood is an extremely bad OOD detector as it obtains AUC-ROC much lower than 0.5 (random guessing), while LR achieves good OOD performances. In addition to AUCROC, we also include Area Under the Curve-Precision Recall Curve (AUCPRC$\uparrow$) and the False Positive Rate at 80\% True Positive Rate (FPR80$\downarrow$) as quantitative measurements.

\subsection{Quantitative Comparison with Other OOD Scores}
After a simple verification of the effectiveness of our proposed OOD score, we carefully study its performances and compare it with competing methods. In our comparison we use likelihood as a simple baseline, and include OOD scores discussed in Section \ref{related_work}, including two variants of input complexity adjusted score (\textbf{IC (png)} and \textbf{IC (jp2)}) \cite{complexity}, likelihood ratio with background model (\textbf{Likelihood Ratio}) \cite{ratio} as well as latent Mahalanobis distance (\textbf{LMD}) \cite{bulusu2020anomalous}. Details of these methods and their implementations can be found in Appendix \ref{competing}. We present two implementations of Likelihood Regret: one optimizes the whole encoder ($\textbf{LR}_{\textbf{E}}$), and the other only optimizes $(\mu_{\mathbf{x}}, \sigma_{\mathbf{x}})$ ($\textbf{LR}_{\textbf{Z}}$).

\begin{table}[H]   
\centering  
\small
\begin{subtable}{.95\textwidth}
\centering
  \begin{tabular}{cccccccc}
   \toprule
   & $\text{LR}_{\text{E}}$ & $\text{LR}_{\text{Z}}$ &  Likelihood & IC (png) & IC (jp2) &Likelihood Ratio & LMD \\
   \midrule
  MNIST & \textbf{0.988} & 0.967 & 0.201 & 0.946 & 0.553 & 0.924 & 0.877 \\
  CIFAR-10 & 0.997 & 0.998 & \textbf{1} & 0.907 & 0.999 & 0.968 & 0.995\\
  SVHN & \textbf{1} & \textbf{1} & 0.999& 0.992 & \textbf{1} & 0.785 & 0.995\\
  KMNIST & \textbf{0.994} & 0.983 & 0.731 & 0.708& 0.599& 0.983 & 0.955\\
  NotMNIST & 0.999 & \textbf{1} & 0.943 & 0.923& 0.966 & 0.996 & 0.998\\
  Noise & \textbf{1} & 0.963 & \textbf{1}& 0.453 & \textbf{1} & \textbf{1} &\textbf{1} \\
  Constant & \textbf{1} & \textbf{1} & 0.928 & \textbf{1}& \textbf{1} &0.775 & 0.981\\
   \bottomrule
   \end{tabular}
\caption{VAE trained on Fashion MNIST}
\end{subtable}

\begin{subtable}{.95\textwidth}
\centering 
  
   \begin{tabular}{cccccccc}
   \toprule
   & $\text{LR}_{\text{E}}$ & $\text{LR}_{\text{Z}}$ &  Likelihood & IC (png) & IC (jp2) &Likelihood Ratio & LMD\\
   \midrule
   MNIST & \textbf{0.998} & 0.976 & 0.008 & 0.994& 0.988 & 0.792 & 0.027 \\
   FMNIST & 0.991 & 0.963 & 0.074 & \textbf{0.992} & 0.990 & 0.807 & 0.183 \\
   SVHN & 0.875 & 0.843 & 0.193 & \textbf{0.912} & 0.908 & 0.265 & 0.279 \\
   LSUN & \textbf{0.691} & 0.640 & 0.494 & 0.624 & 0.315 & 0.632 & 0.527\\
   CelebA & \textbf{0.714} & 0.690 & 0.465 &0.641 &0.564 & 0.447 & 0.576\\
   Noise & 0.994 & 0.922 & \textbf{1} & 0.032 & 0.054 & \textbf{1} & 0.983\\
   Constant & 0.974 & 0.924 & 0.258 & \textbf{1} & \textbf{1} &0.470 & 0.431\\
   \bottomrule
   \end{tabular}
\caption{VAE trained on CIFAR-10}
\end{subtable}
\caption{AUCROC of Likelihood Regret (LR) and other OOD detection scores on different datasets. Each row contains the results of an OOD dataset.} \label{table:main}
\end{table}

The main results of this work are presented in Table \ref{table:main}, which shows the AUCROC scores of different methods on different datasets. We also present results of AUPRC and FPR80 in Table \ref{auprc} and Table \ref{fpr80} in Appendix \ref{additional table}. We see that OOD detection scores exhibit largely consistent behavior across these metrics. We make several important observations from these results:
\begin{enumerate}
  \item We confirm that likelihood is problematic in OOD detection. For example, VAE trained on CIFAR-10 not only assigns significantly higher test likelihood on SVHN, but also on MNIST, Fashion MNIST and random constant images. The severe likelihood misalignment is reflected in an AUCROC value close to $0$. 
  \item Likelihood Regret successfully corrects the misalignment of likelihood and obtains good OOD detection performances across all tasks, as the AUCROC values are close to the optimal value $1$ on all experiments excepts for CIFAR-10 v.s CelebA and CIFAR-10 v.s LSUN. Note that these are the hard cases as all methods do not perform well. Samples from these datasets have similar texture as samples from CIFAR-10, and therefore it is hard  for generative models trained without class labels to distinguish between them. Further, we observe that optimizing the encoder leads to slightly better performance than optimizing $(\mu_{\mathbf{x}}, \sigma_{\mathbf{x}})$ only. One possible explanation is that, for in distribution samples, optimizing the encoder is more constrained than directly optimizing $z$, which prevents the latent variables from moving too much.
  \item While competing methods also obtain good OOD detection performance on some tasks, all of them exhibit severe issues on certain specific tasks. For example, likelihood ratio with background model fails on the classic task of detecting SVHN from in-distribution CIFAR-10 (AUCROC only $0.28$); both variants of input complexity adjusted likelihood fail almost completely on distinguishing between random uniform noise and  CIFAR-10 (AUCROC close to $0$); latent Mahalanobis distance does not perform well on almost all OOD datasets when CIFAR-10 is the in-distribution dataset. These failure cases suggest that none of these OOD scores can be safely applied, at least for VAE models. In contrast, likelihood regret is shown to be effective on all the tasks.
\end{enumerate}

Overall, based on results from Table \ref{table:bpd1}, \ref{auprc} and \ref{fpr80}, we claim that LR achieves the best OOD detection performance. On those tasks where the performance of LR is not the best it is still very close to the best results. More importantly, it achieves good performance without any failure case, while all competing methods are shown to be ineffective on some experiments. Interestingly, the failure cases of competing methods do not exist in their papers where the scores are computed for Glow and PixelCNN. This is partly explained in Section \ref{necessity}. For example, on average, the difference of BPD returned by VAE is only 1.86 nats between CIFAR-10 and noise images, while for Glow and PixelCNN, the differences are $5.33$ nats and $6.6$ nats, respectively. However, the noise images have much larger complexity measurement than CIFAR-10, and the gap of complexity measurement will override the gap of likelihood, leading the input complexity adjusted score  to make the wrong decision. Indeed, we observe that this happens sometimes, even for flow based models (see Appendix \ref{issue}). As for likelihood ratio with background model, this score heavily relies on the contrast between how well a single pixel is modeled by the main model and the background model. However, VAEs are not designed for modeling each single pixel, and the bottleneck structure will ``smooth out" the background, making the contrast with a background model much less effective. 

In summary, the experiments  provide strong evidence that current state-of-the-art OOD scores for generative models may not be applicable to VAE, while our proposed score achieves good OOD detection results on a variety of tasks. To the best of our knowledge, Likelihood Regret is the only effective OOD score that can correct the likelihood misalignment of VAE. 

\textbf{Optimizing other components of the VAE: }
One key reason for the effectiveness of LR on VAE is that we only optimize the configuration of latent variables for a single test input, which is done by optimizing $\mathbf{z}$ directly or optimizing the encoder parameters. The bottleneck structure at the latent variable provides natural regularization that avoids overfitting on the single example. We argue that optimizing other components of the VAE will be less effective, as the optimization can easily overfit any test sample regardless if it is OOD or not. To show this, we perform ablation studies that optimize either the decoder only or the whole VAE (both encoder and decoder), and results are shown in Table \ref{table ablation}. We observe that these alternative approaches are significantly worse than optimizing the latent variables or encoder parameters.

\begin{table}[]  
\centering  
\begin{subtable}{0.45\textwidth}
\small
\centering
\begin{tabular}{ccc}
\toprule
& $\text{LR}_{\text{ED}}$ & $\text{LR}_{\text{D}}$  \\
\midrule
MNIST & 0.124 & 0.189\\
KMNIST & 0.495 & 0.609\\
NotMNIST &0.969 & 0.947\\
Noise & 0.001 & 0.004\\
Constant& 0.989 & 1\\
\bottomrule
\end{tabular}
\caption{VAE trained on Fashion MNIST}
\end{subtable}
\begin{subtable}{0.45\textwidth}
\small
\centering
\begin{tabular}{cccc}
\toprule
& $\text{LR}_{\text{ED}}$ & $\text{LR}_{\text{D}}$ \\
\midrule
SVHN & 0.195 & 0.212\\
LSUN & 0.468 & 0.526\\
CelebA &0.488 & 0.655\\
Noise & 0.002 & 0\\
Constant& 0.322 & 0.435\\
\bottomrule
\end{tabular}
\caption{VAE trained on CIFAR-10}
\end{subtable}
\caption{AUCROC of Likelihood Regret (LR) obtained by optimizing different components of the VAE.$\text{LR}_{\text{ED}}$ corresponds to optimizing both the encoder and decoder, and $\text{LR}_{\text{D}}$ corresponds to optimizing the decoder only.} \label{table ablation}
\end{table}

\textbf{Results on $\beta-$VAE: }VAEs are typically trained with the ELBO objective in \eqref{ELBO}, however, in practice, sometimes we train $\beta-$VAEs, where there is a coefficient $\beta\neq1$ on the KL divergence term. Usually we use $\beta<1$ for better sample quality \citep{dai2019diagnosing,xiao2019generative}, and $\beta>1$ for disentanglement in latent space \citep{higgins2016beta,burgess1804understanding}. We evaluate the OOD detection performances of LR on $\beta-$VAE variants in Table \ref{table beta vae}. From Table \ref{table beta vae}, we observe that LR behaves consistently across different values of $\beta$. 

\begin{table}[]  
\centering  
\begin{subtable}{1\textwidth}
\small
\centering
\begin{tabular}{ccccccccc}
\toprule
& \multicolumn{2}{c}{$\beta = 0.1$} & \multicolumn{2}{c}{$\beta = 0.5$} & \multicolumn{2}{c}{$\beta = 5$} & \multicolumn{2}{c}{$\beta = 10$}\\
& Likelihood & $\text{LR}_{\text{E}}$ & Likelihood & $\text{LR}_{\text{E}}$ & Likelihood & $\text{LR}_{\text{E}}$ & Likelihood & $\text{LR}_{\text{E}}$\\
\midrule
MNIST & 0.191 &	0.988	& 0.154	& 0.948	& 0.231	&0.997	& 0.225 & 0.966\\
KMNIST & 0.704 & 0.997	& 0.679	&0.985&	0.696&	0.996	&0.698&	0.952\\
NotMNIST &0.986	&1&	0.983&	1&	0.985&	1&	0.99&	0.999\\
Noise & 1 &0.989&1&	0.992&	1&	0.996&	1&	0.991\\
Constant& 0.982 &	0.995	&0.964&	0.998&	0.928&	0.998&	0.968&	0.986\\
\bottomrule
\end{tabular}
\caption{$\beta-$VAEs trained on Fashion MNIST}
\end{subtable}
\begin{subtable}{1\textwidth}
\small
\centering
\begin{tabular}{ccccccccc}
\toprule
& \multicolumn{2}{c}{$\beta = 0.1$} & \multicolumn{2}{c}{$\beta = 0.5$} & \multicolumn{2}{c}{$\beta = 5$} & \multicolumn{2}{c}{$\beta = 10$}\\
& Likelihood & $\text{LR}_{\text{E}}$ & Likelihood & $\text{LR}_{\text{E}}$ & Likelihood & $\text{LR}_{\text{E}}$ & Likelihood & $\text{LR}_{\text{E}}$\\
\midrule
SVHN & 0.189 &	0.847	&0.171&	0.835&	0.194&	0.866&	0.153&	0.873\\
LSUN & 0.474 &	0.645	&0.465&	0.655&	0.469	&0.654	&0.442	&0.657\\
CelebA &0.499 &	0.709&	0.477&	0.714&	0.528&	0.718&	0.462&	0.706\\
Noise & 1&	1&	1&	0.999&	1&	0.994&	1&	0.9735\\
Constant& 0.319	&0.964&	0.279&	0.947&	0.268&	0.962&	0.284&	0.954\\
\bottomrule
\end{tabular}
\caption{$\beta-$VAEs trained on CIFAR-10}
\end{subtable}
\caption{AUCROC of Likelihood Regret (LR) and Likelihoods for $\beta-$VAEs on different datasets.} \label{table beta vae}
\end{table}

\textbf{On the capacity of the VAEs: }One concern regarding the effectiveness of Likelihood Regret is that it may depend on the capacity of VAEs. For example, a VAE with large capacity is easier to optimize for the encoder configuration on a single sample, leading to larger Likelihood Regret score. we evaluate LR on VAEs with different capacities, and results are presented in Appendix \ref{capacity section}, Table \ref{table capacity}. We conclude that the performance of LR is robust to the capacity of VAEs. The AUCROC slightly drops for VAEs with large capacity trained on CIFAR-10. By inspection, we observe that large VAEs overfit the training data, as the test NLL on CIFAR-10 is even larger than that of the baseline VAE. In the case of overfitting, the LR for in-distribution test data will be larger. 

\textbf{Runtime: }Our method requires $2$ likelihood estimations and several optimization iterations for each testing image, which will make it slower than its competing methods. As a comparison, input complexity adjusted likelihood does not have computationally overhead, and likelihood ratio with background model needs to train $2$ models and take $2$ likelihood estimations. However, we observe that in our experiments, on average the computation of LR takes less than $0.3$s, which is comparable to the IWAE likelihood estimation (also around $0.3$s), so the computational overhead is acceptable. 

\textbf{Additional Results: }In Appendix \ref{app_reverse}, we include OOD results of models trained on MNIST and SVHN for completeness. We observe that our method works well in these experiments, while competing OOD scores still exhibit severe issues. In appendix \ref{illu}, we show some examples of reconstruction before and after the optimization of encoder. In Appendix \ref{app_recon}, we show more qualitative examples of reconstruction on different test datasets. In Appendix  \ref{app_sample}, we display some randomly generated samples from the VAEs.
\vspace{-3mm}
\section{Conclusion}
In this paper, we carefully study the task of unsupervised out-of-distribution detection for VAEs. We evaluate some current state-of-the-art OOD detection scores on a set of experiments, and we conclude that their success on OOD detection for other probabilistic models cannot be easily transferred to VAEs. We also try to provide clues to show that OOD detection is harder for VAEs than for other generative models. To overcome the difficulty, we propose Likelihood Regret, an OOD score for VAEs that is effective on all the tasks we evaluated. We believe that Likelihood Regret can be extended to other generative models if a good optimizable model configuration is defined. We hope this work can lead to further progress on OOD detection for probabilistic generative models.

\newpage
\section*{Broader Impact}
Out-of-distribution detection is a research direction with significant social impact. Nowadays, deep learning is deployed on many systems that are making critical decisions, such as medical diagnosis, factory manufacturing, autonomous driving. Being able to detect anomalous cases is crucial for these systems. Therefore, our work, together with many other research efforts on out-of-distribution detection, is very important for the future development of artificial intelligence. However, we should be cautious on solely relying on algorithmic anomaly detection, as there is always the risk of certain anomalous cases that can fool the algorithms. It is very risky to completely trust these imperfect algorithms.

\bibliographystyle{plainnat}
\bibliography{main}
\newpage
\appendix
\section{Experimental Settings}\label{appA}
In this section, we introduce detailed settings of our experiments. 
\subsection{Datasets} \label{datasets}
\begin{figure}[ht]
    \centering
    \begin{subfigure}{.45\linewidth}
        \includegraphics[scale=0.55]{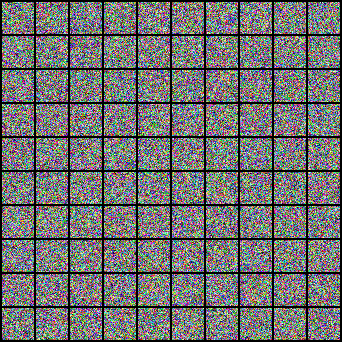}
        \caption{Noise}
    \end{subfigure}
    \hskip2em
    \begin{subfigure}{.45\linewidth}
        \includegraphics[scale=0.55]{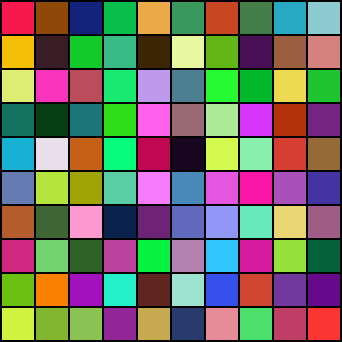}
        \caption{Constant}
    \end{subfigure}
    \hskip2em
    \begin{subfigure}{.45\linewidth}
        \includegraphics[scale=0.55]{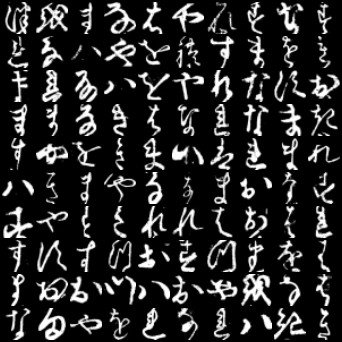}
        \caption{KMNIST}
    \end{subfigure}
    \hskip2em
   \begin{subfigure}{.45\linewidth}
        \includegraphics[scale=0.55]{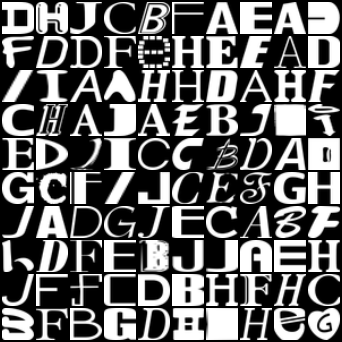}
        \caption{NotMNIST}
    \end{subfigure}
    \caption{\label{app_ownimage}
     Some examples of our created images used in experiments.}
\end{figure}

We use several publicly available datasets in our experiments. These datasets include MNIST \cite{MNIST}, Fashion-MNIST\cite{fmnist}, KMNIST\cite{kmnist}, notMNIST, CIFAR-10 and CIFAR-100\cite{cifar}, SVHN \cite{svhn}, CelebA \cite{celeba} and LSUN\cite{yu2015lsun}. We also create two types of synthetic images: Noise and Constant. Noise images are random samples from $\text{Unif}\{0,...,255\}$ distribution for each pixel, and Constant images are images with constant value sampled from $\text{Unif}\{0,...,255\}$ for each channel. Some examples of our created images are shown in Figure \ref{app_ownimage} We also present some examples of KMNIST and NotMNIST, as they are less familiar to the public. They are used as OOD datasets for models trained on Fashion MNIST. 

For most of our experiments, the VAE is trained on Fashion-MNIST and CIFAR-10, where we use the training partition of the datasets. For other datasets used for testing, we use a test partition if it is available, and use randomly sampled data if no predefined partition is available.

We resize images to spatial dimension $32 \times 32$ for all datasets. We trained VAE on Fashion MNIST using $32 \times 32 \times 1$ images, and when we test it on color images, we use only the first channel of the color image. When we use Fashion MNIST or MNIST images to test the VAE trained on CIFAR, we copy the channel three times to make them $32 \times 32 \times 3$. 

When computing quantitative metrics, we use $5000$ randomly chosen images from in-distribution and OOD datasets, respectively.

\subsection{Implementation Detail}\label{implement}
The training and testing of our models largely follow the setting of \cite{nalisnick}. In particular, we train VAEs with the DCGAN \cite{dcgan} structure. We present the network structure in Table \ref{netstruct}. 

\begin{table}
  \centering
  \begin{tabular}{ll}
    \toprule
    Encoder   & Decoder    \\
    \midrule
     Input $x$   &      Input $z$, reshape to $\text{nz} \times 1 \times 1$ \\
     $4 \times 4$ $\text{Conv}_{\text{nf}}$ Stride $2$, BN, ReLU   &  $4 \times 4$ $\text{Deconv}_{4 \times \text{nf}}$ Stride $1$, BN, ReLU\\
     
     $4 \times 4$ $\text{Conv}_{2 \times \text{nf}}$ Stride $2$, BN, ReLU & $4 \times 4$ $\text{Deconv}_{2 \times \text{nf}}$ Stride $2$, BN, ReLU  \\
     
     $4 \times 4$ $\text{Conv}_{4 \times \text{nf}}$ Stride $2$, BN, ReLU & $4 \times 4$ $\text{Deconv}_{\text{nf}}$ Stride $2$, BN, ReLU \\
     
     $4 \times 4$ $\text{Conv}_{2 \times \text{nz}}$ Stride $1$ & $4 \times 4$ $\text{Conv}_{256\times\text{nc}}$ Stride $2$\\
    \bottomrule
  \end{tabular}
   \caption{Network structure for VAE based on DCGAN. $\text{nz} = 100$ for all models. For VAE trained on Fashion MNIST, $\text{nf}=32, \text{nc} = 1$; for VAE trained on CIFAR-10, $\text{nf}=64, \text{nc} = 3$.}\label{netstruct}
\end{table}

On Fashion MNIST we train the VAE for 100 epochs with constant learning rate $5e-4$ using Adam optimizer and batch size $64$. On CIFAR-10 we train the VAE for 200 epochs with constant learning rate $5e-4$ using Adam optimizer and batch size $64$. When computing Likelihood Regret, we have the choice of optimizing the whole encoder or only optimizing the mean and variance of posterior. For the former, we start with the trained encoder and optimize its parameters for $100$ steps using the Adam optimizer with learning rate $1e-4$. For the later, we start with the encoding mean and variance, and run optimization for 300 steps using Adam optimizer with learning rate $1e-4$. 

\subsection{Implementing Competing Methods}\label{competing}
Input complexity adjusted likelihood \cite{complexity} is computed by subtracting a measure of the input's complexity from the negative log likelihood with. The input complexity can be obtained from the length of the binary string returned by some lossless compression algorithms. We simply follow their work and use PNG compression and JPEG2000 compression implemented in OpenCV.  

Likelihood ratio with background model \cite{ratio} is computed by subtracting the log likelihood of the background model from the log likelihood of the main model. The background model is trained by perturbing a proportion of randomly chosen pixels, where the perturbation is done by replacing the pixel value by a uniformly sampled random value between 0 and 255. One of the key hyper-parameters $\mu$, is the  percentage of pixels to be perturbed. The authors suggest to choose $\mu$ between $0.1$ and $0.3$. We do a simple grid search on $\{0.1,0.2,0.3\}$, and use the best one ($0.3$ for Fashion MNIST, and $0.2$ for CIFAR-10). We trained the background VAE with the same setting of the main VAE, except that we apply $\lambda = 10$ $L_2$ weight decay as suggested by the authors. 

Latent Mahalanobis distance \cite{bulusu2020anomalous} combines the reconstruction loss and Mahalanobis distance in the latent space as an OOD score. In particular, their score is defined by 
\begin{align*}
    \text { novelty }(\mathbf{x})=\alpha \cdot D_{M}(E(\mathbf{x}))+\beta \cdot \ell(x, D(E(\mathbf{x}))),
\end{align*}
where $D$ and $E$ are the decoder and encoder, respectively. The second term is the reconstruction loss, and the first term is the Mahalanobis distance ($D_{M}$) between the
encoded sample and the mean vector of the encoded training set. Their method is designed for auto-encoders, so we take $E(x)$ to be the mean of the variational posterior output by the encoder. As suggested by the author, to balance the two terms, $\alpha$ was set to the reciprocal of the standard deviation of the Mahalanobis distance between the encoded validation data and the mean latent train vector, and $\beta$ to the reciprocal of the standard deviation of the reconstruction error on the validation set.

\section{Additional Quantitative results: AUPRC and FPR80}\label{additional table}

\begin{table}[H]   
\centering  
\begin{subtable}{.95\textwidth}
\centering
  \begin{tabular}{cccccccc}
   \toprule
   & $\text{LR}_{\text{E}}$ & $\text{LR}_{\text{Z}}$ &  Likelihood & IC (png) & IC (jp2) &Likelihood Ratio & LMD \\
   \midrule
  MNIST & \textbf{0.980} &	0.938 &	0.344&		0.923	& 0.563 & 0.917	& 0.866 \\
  CIFAR-10 & 0.995 & 0.998	& \textbf{1}	& 0.904	& \textbf{1}		 &0.912	& 0.998\\
  SVHN & 0.998	& \textbf{1} & 0.996	& 0.993 & 1 & 0.621 & 0.999\\
  KMNIST & \textbf{0.993} & 0.985 & 0.734 & 0.642	& 0.568	& 0.984	&	0.962\\
  NotMNIST & 0.999 & \textbf{1}	 & 0.935	& 0.943 & 0.953	& 0.996 &	0.999\\
  Noise & 0.983 & 0.954& \textbf{1}	&	0.4342 &	\textbf{1}	&	\textbf{1}	&	\textbf{1}
 \\
  Constant & 0.999 & \textbf{1}&	0.915 & \textbf{1}	& \textbf{1}	& 0.628 & 0.985\\
   \bottomrule
   \end{tabular}
\caption{VAE trained on Fashion MNIST}
\end{subtable}
\begin{subtable}{.95\textwidth}
\centering 
  
   \begin{tabular}{cccccccc}
   \toprule
   & $\text{LR}_{\text{E}}$ & $\text{LR}_{\text{Z}}$ &  Likelihood & IC (png) & IC (jp2) &Likelihood Ratio & LMD\\
   \midrule
   MNIST & 0.993& 0.979 &	0.307		& \textbf{0.997} & 0.988		& 0.665	& 0.308 \\
   FMNIST & 0.980 & 0.956 &	0.314 	& 0.982 & \textbf{0.987} &	0.656 &		0.337\\
   SVHN & 0.841	& 0.799 & 0.343 &		\textbf{0.922}	& 0.914 &		0.337  & 0.389\\
   LSUN & \textbf{0.681} & 0.610	& 0.508	& 0.611	& 0.408	& 0.611	& 0.520\\
   CelebA & \textbf{0.715} & 0.634 & 0.474		& 0.572	& 0.509	 &0.405 & 0.524\\
   Noise & 0.940& 0.814	& \textbf{1}	& 0.313&	0.318&	\textbf{1} & 0.964\\
   Constant & 0.965 &	0.924	& 0.445 & \textbf{1} &	\textbf{1}		& 0.410	 & 0.593\\
   \bottomrule
   \end{tabular}
\caption{VAE trained on CIFAR-10}
\end{subtable}
\caption{AUCPRC of Likelihood Regret (LR) and other OOD detection scores on different datasets.} \label{auprc}
\end{table}

\begin{table}[H]   
\centering  
\begin{subtable}{.95\textwidth}
\centering
  \begin{tabular}{cccccccc}
   \toprule
   & $\text{LR}_{\text{E}}$ & $\text{LR}_{\text{Z}}$ &  Likelihood & IC (png) & IC (jp2) &Likelihood Ratio & LMD \\
   \midrule
  MNIST & 0.02 & 0.09 & 0.97	& 0.1 & 0.68 & 0.15	& 0.2\\
  CIFAR-10 & 0 &0 &	0 & 0.16 & 0	&	0.02 & 0\\
  SVHN & 0	&0 & 0 & 0 & 0 & 0.35	&0\\
  KMNIST & 0.01 &	0.01	& 0.51		& 0.44	& 0.57	& 0.02	& 0.06 \\
  NotMNIST & 0 &0 &0.01	& 0.11	&0.01		& 0.01	&0\\
  Noise&  0	& 0.06	& 0	& 0.57	& 0		& 0	&0 \\
  Constant &0	& 0	& 0.06	&0	&0		& 0.27	&0.01\\
   \bottomrule
   \end{tabular}
\caption{VAE trained on Fashion MNIST}
\end{subtable}
\begin{subtable}{.95\textwidth}
\centering 
  
   \begin{tabular}{cccccccc}
   \toprule
   & $\text{LR}_{\text{E}}$ & $\text{LR}_{\text{Z}}$ &  Likelihood & IC (png) & IC (jp2) &Likelihood Ratio & LMD\\
   \midrule
   MNIST & 0.01 &	0.02	&1		& 0.01& 0	&0.27	&1\\
   FMNIST & 0.02 & 0.06	& 0.99		& 0.01 & 0.01	& 0.32	&0.98\\
   SVHN &0.18 & 0.27 & 0.97	& 0.05	&0.1 & 0.91	& 0.97\\
   LSUN & 0.46 & 0.55& 0.77		&0.5&	0.86 & 0.62	& 0.77\\
   CelebA & 0.37 & 0.48 &	0.68&		0.55 & 0.67	& 0.82	& 0.67\\
   Noise & 0.02 & 0.08	& 0	& 0.99	& 0.93 & 0 & 0\\
   Constant & 0.03 & 0.09 & 0.98	&	0 & 	0	& 0.65	& 0.96\\
   \bottomrule
   \end{tabular}
\caption{VAE trained on CIFAR-10}
\end{subtable}
\caption{FPR80 of Likelihood Regret (LR) and other OOD detection scores on different datasets.} \label{fpr80}
\end{table}

\section{Results of Model Trained on SVHN}\label{app_reverse}
In Table \ref{svhn_table}, we include results of a set of simple experiments for VAE trained on SVHN. Note that in this case, likelihood itself works well on most tasks. We do not show results of VAE trained on MNIST because every method works nearly perfectly. We observe that LR achieves good performances on all the tasks, while input complexity still has trouble with distinguishing noise from in-distribution data. In addition, its performance on SVHN v.s. CIFAR-10 lies far behind LR. As in Table \ref{table:main}, likelihood ratio has trouble on CIFAR-10 v.s. Constant. These failures suggest that competing OOD scores have systematical issues on VAE. 

\begin{table}[H]   
\centering  
\centering
  \begin{tabular}{cccccccc}
   \toprule
   & $\text{LR}_{\text{E}}$ & $\text{LR}_{\text{Z}}$ &  Likelihood & IC (png) & IC (jp2) &Likelihood Ratio & LMD \\
   \midrule
  MNIST & \textbf{1} &\textbf{1} & \textbf{1}  & \textbf{1} & \textbf{1} & \textbf{1} & \textbf{1}\\
  FMNIST & \textbf{1} &0.999 & \textbf{1} & \textbf{1} & \textbf{1} & 0.928 &0.998\\
  CIFAR-10 & 0.924 & 0.842 & \textbf{0.982}  & 0.524 & 0.608 & 0.936 & 0.955 \\
  Noise & \textbf{1} & \textbf{1}& \textbf{1}& 0.235 & 0.105 & \textbf{1} & \textbf{1}\\
  Constant & \textbf{1} & \textbf{1} & 0.213 & \textbf{1}& \textbf{1} & 0.136 & 0.818 \\
   \bottomrule
   \end{tabular}
\caption{AUCROC for model trained on SVHN}\label{svhn_table}
\end{table}

\section{Issues with Input Complexity on Glow}\label{issue}
In this section, we show the OOD results of input complexity adjusted likelihood for Glow trained on SVHN and MNIST. Note that this set of experiments are not performed by the authors of \cite{complexity}. We use the same Glow structure used in \cite{complexity}, and train Glow on SVHN and MNIST dataset. We use input complexity adjusted likelihood to detect OOD samples from Fashion MNIST and CIFAR-10. Basically this is the reverse of the commonly conducted experiments. The AUCROC of MNIST v.s. Fashion MNIST is $0.633$, and the AUCROC of SVHN v.s. CIFAR-10 is $0.518$. Both results suggest that input complexity adjusted likelihood may not work in general, even on flow based model.
\section{Illustration of Optimization} \label{illu}
\begin{figure}[ht]
    \centering
     \begin{subfigure}{.45\linewidth}
        \includegraphics[scale=0.5]{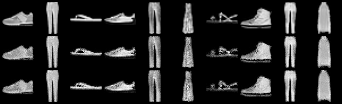}
        \caption{Fashion MNIST}
    \end{subfigure}
    \hskip2em
    \begin{subfigure}{.45\linewidth}
        \includegraphics[scale=0.5]{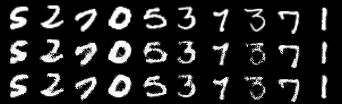}
        \caption{MNIST}
    \end{subfigure}
    \hskip2em
    \begin{subfigure}{.45\linewidth}
        \includegraphics[scale=0.5]{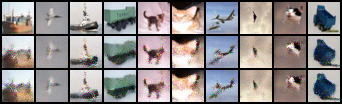}
        \caption{CIFAR-10}
    \end{subfigure}
    \hskip2em
   \begin{subfigure}{.45\linewidth}
        \includegraphics[scale=0.5]{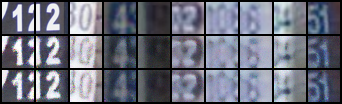}
        \caption{SVHN}
    \end{subfigure}
    \caption{\label{fig:recon_compare}
     For each subfigure, the top row contains original images, the middle row contains the reconstruction from VAE, and the bottom row contains the reconstruction images with optimized encoder. \textbf{(a), (b)} are obtained from VAE trained on Fsahion MNIST. \textbf{(c), (d)} are obtained from VAE trained on CIFAR-10.}
\end{figure}

To better illustrate Likelihood Regret, we compare the reconstruction of some test samples with trained VAE and optimized posterior distribution in Figure \ref{fig:recon_compare}. Since as we show in Figure \ref{fig:recon}, the VAE can reconstruct both the in-distribution and out of distribution data very well, visually we cannot see obvious differences between reconstructed images from VAE and from optimized $q_{\phi}(\mathbf{z}|\mathbf{x})$. However, we can still observe the improvements of reconstruction by the optimal $q_{\phi}(\mathbf{z}|\mathbf{x})$ in MNIST examples.

\section{More Reconstruction Examples}\label{app_recon}
\begin{figure}[ht]
    \centering
     \begin{subfigure}{.45\linewidth}
        \includegraphics[scale=0.5]{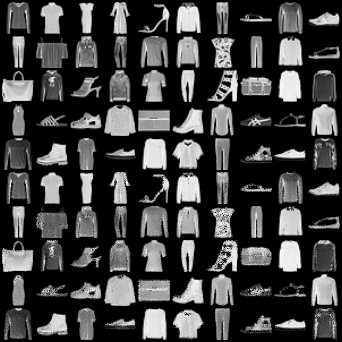}
        \caption{Fashion MNIST}
    \end{subfigure}
    \hskip2em
    \begin{subfigure}{.45\linewidth}
        \includegraphics[scale=0.5]{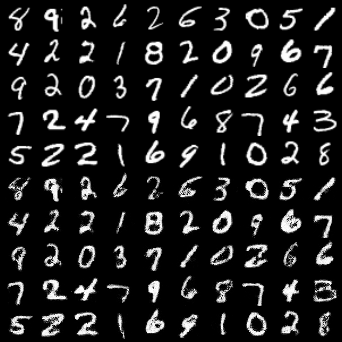}
        \caption{MNIST}
    \end{subfigure}
    \hskip2em
    \begin{subfigure}{.45\linewidth}
        \includegraphics[scale=0.5]{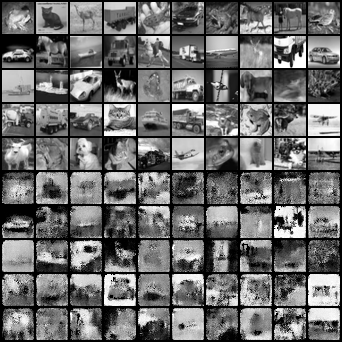}
        \caption{CIFAR-10}
    \end{subfigure}
    \hskip2em
   \begin{subfigure}{.45\linewidth}
        \includegraphics[scale=0.5]{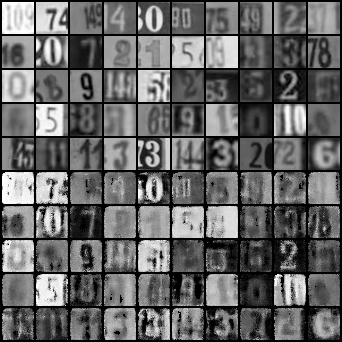}
        \caption{SVHN}
    \end{subfigure}
    \caption{\label{fig:recon_large1}
     Some examples of reconstructed images using VAE trained on Fashion MNIST. For each subfigure, the first 5 rows are original images and the last 5 rows are their corresponding reconstructions.}
\end{figure}

\begin{figure}[ht]
    \centering
     \begin{subfigure}{.45\linewidth}
        \includegraphics[scale=0.58]{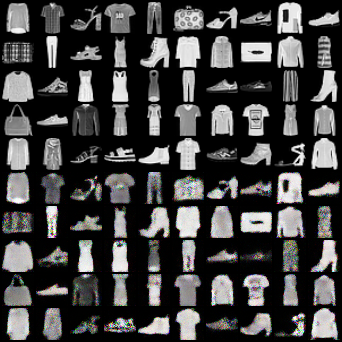}
        \caption{Fashion MNIST}
    \end{subfigure}
    \hskip2em
    \begin{subfigure}{.45\linewidth}
        \includegraphics[scale=0.58]{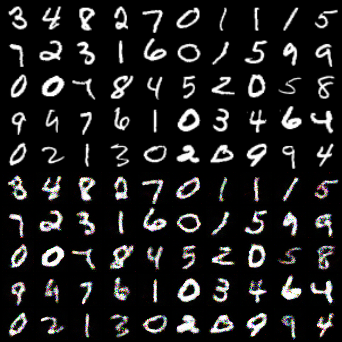}
        \caption{MNIST}
    \end{subfigure}
    \hskip2em
    \begin{subfigure}{.45\linewidth}
        \includegraphics[scale=0.58]{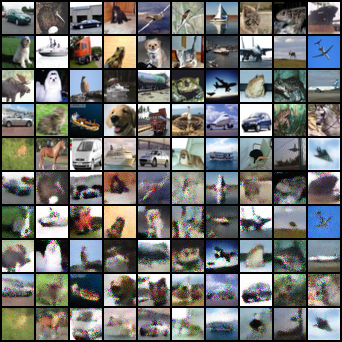}
        \caption{CIFAR-10}
    \end{subfigure}
    \hskip2em
   \begin{subfigure}{.45\linewidth}
        \includegraphics[scale=0.58]{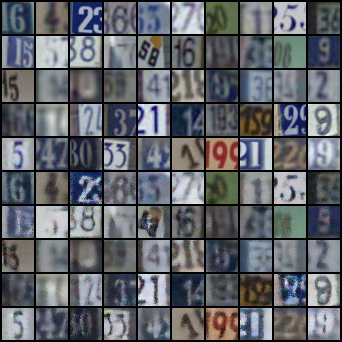}
        \caption{SVHN}
    \end{subfigure}
    \caption{\label{fig:recon_large2}
     Some examples of reconstructed images using VAE trained on CIFAR-10. For each subfigure, the first 5 rows are original images and the last 5 rows are their corresponding reconstructions.}
\end{figure}

In this section, we present examples of reconstructed images in different datasets. See Figure \ref{fig:recon_large1} and Figure \ref{fig:recon_large2}. We observe that VAE trained on CIFAR-10 can reconstruct images from multiple datasets very well. 

\section{Robustness of LR w.r.t the Capacity of VAEs} \label{capacity section}
We present results on VAEs with different capacity in Table \ref{table capacity}.
\begin{table}[]  
\centering  
\begin{subtable}{1\textwidth}
\small
\centering
\begin{tabular}{ccccccccc}
\toprule
& \multicolumn{2}{c}{$C = \frac{1}{4}\times$} & \multicolumn{2}{c}{$C = \frac{1}{2}\times$} & \multicolumn{2}{c}{$C = 2\times$} & \multicolumn{2}{c}{$C = 4\times$}\\
& Likelihood & $\text{LR}_{\text{E}}$ & Likelihood & $\text{LR}_{\text{E}}$ & Likelihood & $\text{LR}_{\text{E}}$ & Likelihood & $\text{LR}_{\text{E}}$\\
\midrule
MNIST & 0.110&	0.966	&	0.148	&0.986&		0.125	&0.975&		0.237&	0.984\\
KMNIST & 0.608&	0.991	&	0.65	&0.996&		0.711&	0.992&		0.812&	0.994\\
NotMNIST &0.977	&1	&	0.978	&1	&	0.99	&0.999&		1	&0.997\\
Noise & 1	&0.987	&	1	&0.999&		1	&0.997&		1	&1\\
Constant& 0.966 &	0.998&		0.947&	0.998&		0.957&	0.994&		0.979	&0.997\\
\bottomrule
\end{tabular}
\caption{$\beta-$VAEs trained on Fashion MNIST}
\end{subtable}
\begin{subtable}{1\textwidth}
\small
\centering
\begin{tabular}{ccccccccc}
\toprule
& \multicolumn{2}{c}{$C = \frac{1}{4}\times$} & \multicolumn{2}{c}{$C = \frac{1}{2}\times$} & \multicolumn{2}{c}{$C = 2\times$} & \multicolumn{2}{c}{$C = 4\times$}\\
& Likelihood & $\text{LR}_{\text{E}}$ & Likelihood & $\text{LR}_{\text{E}}$ & Likelihood & $\text{LR}_{\text{E}}$ & Likelihood & $\text{LR}_{\text{E}}$\\
\midrule
SVHN & 0.198 &	0.868	&	0.176&	0.875	&	0.203	&0.814	&	0.198	&0.79\\
LSUN & 0.462&	0.641		&0.453	&0.652&		0.445&	0.635&		0.467&	0.611\\
CelebA &0.455	&0.706	&	0.445	&0.717&		0.521&	0.793&		0.475&	0.753\\
Noise & 1	&1	&	1	&1&		1	&0.998&		1	&0.999\\
Constant& 0.218	&0.999&		0.276&	0.998&		0.292&	0.989&		0.246	&0.962\\
\bottomrule
\end{tabular}
\caption{$\beta-$VAEs trained on CIFAR-10}
\end{subtable}
\caption{AUCROC of Likelihood Regret (LR) and Likelihoods for VAEs with different capacity, where we proportionally increase/decrease the number of channels of convolution layers. For example, $C = \frac{1}{4}\times$ means that the VAE has $\frac{1}{4}$ channels compared to the baseline VAE.} \label{table capacity}
\end{table}

\section{Randomly Generated Samples}\label{app_sample}
\begin{figure}[ht]
    \centering
     \begin{subfigure}{.45\linewidth}
        \includegraphics[scale=0.58]{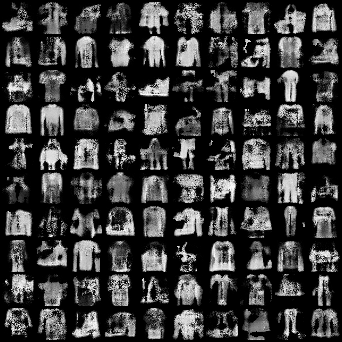}
        \caption{Fashion MNIST}
    \end{subfigure}
    \hskip2em
    \begin{subfigure}{.45\linewidth}
        \includegraphics[scale=0.58]{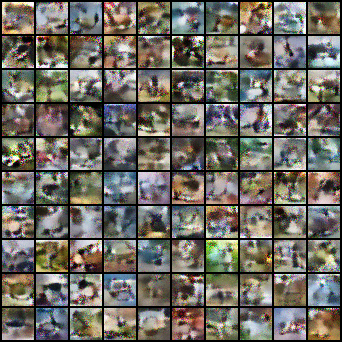}
        \caption{CIFAR}
    \end{subfigure}
    \caption{\label{fig:sample}
     Some examples of randomly generated samples.}
\end{figure}

We show some randomly generated samples from VAEs in Figure \ref{fig:sample}. Although the samples are blurry and a bit noisy (due to the cross entropy loss), the semantics of the samples suggest that our VAEs model the in-distribution data well.
\end{document}